%% file: MorphoFlow.tex
\documentclass[runningheads]{llncs}
\usepackage{xcolor}
\usepackage{graphicx}
\usepackage{multirow}
\usepackage{booktabs}
\usepackage{hyperref}
\usepackage{amsmath}
\usepackage[T1]{fontenc}
\usepackage{amssymb}
\usepackage{xspace}
\usepackage{rotating}
\newcommand{\name}[0]{{\sc MorphoFlow}\xspace}
\usepackage{graphicx,verbatim}
\usepackage{amsmath}
\begin{document}
%
%\title{\name: Generative and Compact Statistical Shape Modeling from Sparse Surface Annotations Using sparsity inducing Priors}
%\title{\name: Generative Shape Modeling with Adaptive Latent Relevance from Sparse Annotations}
\title{\name: Sparse-Supervised Generative Shape Modeling with Adaptive Latent Relevance}
% \title{\name: Generative Shape Modeling with Adaptive Latent Relevance Under Sparse Supervision}
\titlerunning{\name: Sparse-Supervised Generative Shape Modeling}
% If the paper title is too long for the running head, you can set
% an abbreviated paper title here
%
% \begin{comment}  %% Removed for anonymized MICCAI submission
\author{Mokshagna Sai Teja Karanam \and
Tushar Kataria \and
Shireen Elhabian}
\authorrunning{M. Karanam et al.}
% First names are abbreviated in the running head.
% If there are more than two authors, 'et al.' is used.

\institute{Kahlert School of Computing, 
Scientific Computing and Imaging Institute, \\University of Utah, Utah, USA \\
\email{\{u1418261@,tushar.kataria@,shireen@sci.\}utah.edu}}

% \end{comment}

%\author{Anonymized Authors}  %% Added for anonymized MICCAI submission
%\authorrunning{Anonymized Author et al.}
%\institute{Anonymized Affiliations \\
%    \email{email@anonymized.com}}
  
\maketitle              % typeset the header of the contribution
\begin{abstract}
Statistical shape modeling (SSM) is central to population-level analysis of anatomical variability, yet most existing approaches rely on densely annotated segmentations and fixed latent representations. These requirements limit scalability and reduce flexibility when modeling complex anatomical variation. 
We introduce \name, a sparse-supervised generative shape modeling framework that learns compact probabilistic shape representations directly from sparse surface annotations.
\name integrates neural implicit shape representations with an autodecoder formulation and autoregressive normalizing flows to learn an expressive probabilistic density over the latent shape space.
The neural implicit representation enables resolution-agnostic modeling of 3D anatomy, while the autodecoder formulation supports direct optimization of per-instance latent codes under sparse supervision.
The autoregressive flow captures the distribution of latent anatomical variability, providing a tractable, likelihood-based generative model of shapes.
To promote compact and structured latent representations, we incorporate adaptive latent relevance weighting through sparsity-inducing priors, enabling the model to regulate the contribution of individual latent dimensions according to their relevance to the underlying anatomical variation while preserving generative expressivity.
The resulting latent space supports uncertainty quantification and anatomically plausible shape synthesis without manual latent dimensionality tuning.
Evaluation on publicly available lumbar vertebrae and femur datasets demonstrates accurate high-resolution reconstruction from sparse inputs and recovery of structured modes of anatomical variation consistent with population-level trends. \href{Code will be released Upon Acceptance}{Code will be released upon acceptance}.

% Authors must provide keywords and are not allowed to remove this Keyword section.

\end{abstract}
\input{Introduction}

% \begin{itemize}

% \item Proposed an autoregressive normalizing flow prior with ARD regularization to obtain a compact and structured latent representation that can match the underlying shape population. 

% \item Resulting in generative latent formulation enabling anatomically plausible sampling without compromising reconstruction quality. Analyzing modes of variations of the highest mode according to ard.

% \item Quantitative evaluation including uncertaining and surface reconstruction errors due to underlying of slices at inference. Results are shown on publicly available femur and lumbar datasets. 
% %, including downstream femur pathology classification.

% \end{itemize}

\input{Methods}

\input{Results}

\vspace{-0.5em}
\section{Conclusion and Future Work}
\vspace{-0.5em}
We presented \name, a sparsely-supervised generative framework for statistical shape modeling that combines neural implicit auto-decoders with an auto-regressive normalizing flow prior and adaptive latent relevance weighting. The proposed approach learns a compact, structured probabilistic latent space directly from sparse surface annotations while preserving high-fidelity, resolution-agnostic reconstruction.
Our results show that incorporating the Flow and ARD prior preserves reconstruction accuracy across latent dimensionalities while promoting compact and interpretable representations. The learned density aligns geometric distances in shape space with the likelihood structure of the latent space, enabling principled sampling and anatomically plausible shape synthesis. Furthermore, the generative formulation facilitates uncertainty quantification of the occupancy model, improving interpretability 
%and helping identify failure modes, such as elevated uncertainty in high-curvature regions.
%
Future work will focus on exploring ordinal ARD formulations that explicitly rank latent dimensions according to dominant modes of variation, %developing improved surface reconstruction models to enhance accuracy in high-curvature regions, 
and establishing standardized benchmarks for shape modeling under limited-annotation regimes. % using \name.

\begin{credits}

\noindent\textbf{Acknowledgments}\par

\hspace{-1.7em}This work was supported by the National Institutes of Health under grant number R01 DE032366.

\end{credits}
\newpage
\bibliographystyle{splncs04}
\bibliography{MorphoFlow}
\end{document}

%% file: Introduction.tex
\section{Introduction}

Statistical shape models (SSMs) characterize population-level anatomical variability by learning compact representations of shape variation. These models support shape reconstruction \cite{smelkina2017reconstruction}, treatment planning \cite{erdur2025deep}, disease analysis \cite{peiffer2022statistical,khan2022machine}, and their utility depends on faithfully capturing real-world morphological diversity while remaining robust to noise and incomplete data \cite{iyer2025mesh2ssm++,luthi2017gaussian}. 
SSMs seek a latent vector space in which distances correspond to meaningful geometric (and pathological) differences, enabling both quantitative analysis \cite{cates2017shapeworks} and generative modeling \cite{laga2024statistical}. 
In practice, supervision for shape modeling is typically provided by segmentation masks derived from medical images (e.g. CT and MRI) \cite{adams2024weakly,cates2017shapeworks,iyer2025mesh2ssm++}. 
However, in many clinical scenarios, such supervision is not easy to obtain requiring medical professionals to costly and labor-intensive, requiring medical professionals to manually delineate the organ of interest, thereby substantially increasing the effort required to obtain full volumetric masks \cite{ryabtsev2025streamlining,wong2024scribbleprompt,cheng2025interactive}. 
To reduce this burden, segmentation masks may be available only for a limited number of slices, may be sparsely annotated, or may originate from low-resolution or anisotropic acquisitions with thick slices \cite{stutz2020learning,ukey2024weakly}. 
Furthermore, the imaging process itself limits the observable morphology: coarse resolution constrains geometric detail and thus weakens the supervision signal for statistical modeling \cite{yu2023techniques}. 
Consequently, many practical SSM pipelines operate under weak or incomplete supervision, where only partial geometric information is available to learn population-level shape representations \cite{goceri2019challenges,shen2023survey}.

A well-formed statistical shape model should satisfy three essential properties. 
First, it should embed shapes in a structured latent space where distances reflect morphological variation, enabling meaningful geometric comparisons \cite{luthi2017gaussian,cates2017shapeworks}. 
Second, the latent space should capture anatomically relevant features for downstream tasks such as classification and pathology analysis \cite{bhalodia2024deepssm,adams2024point2ssm++}. 
Third, the model should be generative, enabling synthesis of plausible new shapes and reconstruction beyond the training resolution \cite{rezende2015variational,dinh2016density,park2019deepsdf}. 
Importantly, reconstruction fidelity should not deteriorate under increasing input sparsity, and sub-voxel geometric accuracy should be preserved despite coarse supervision \cite{mescheder2019occupancy}. 
These requirements jointly imply that the latent space must be both geometrically meaningful and probabilistically well-structured, particular when supervision is sparse.
%
%Current optimization- and deep learning-based SSM approaches (e.g., \cite{cates2017shapeworks,bhalodia2024deepssm,adams2024point2ssm++,iyer2023mesh2ssm}) typically rely on densely segmented shapes or high-resolution volumetric data.
%
%Under weak supervision, these methods often struggle to recover fine geometric detail, maintain compact latent structure, or generalize beyond the resolution of the training data \cite{zhang2026weakly,ukey2024weakly}. 
Current optimization- and deep learning-based SSM methods \cite{cates2017shapeworks,bhalodia2024deepssm,adams2024point2ssm++,iyer2023mesh2ssm} rely on densely segmented shapes or high-resolution volumes and, under weak supervision, often struggle to capture fine geometric detail, learn compact latent representations, or generalize beyond the training resolution \cite{zhang2026weakly,ukey2024weakly}.

Recent advances in neural implicit representations provide a principled foundation to address these limitations. Implicit neural fields, such as signed distance and occupancy networks (e.g., \cite{park2019deepsdf,mescheder2019occupancy}), represent shapes as continuous functions rather than discrete meshes or voxel grids \cite{amiranashvili2024learning}. 
By modeling surfaces implicitly, these methods eliminate the need for explicit correspondences and enable evaluation at arbitrary spatial resolution. 
This continuous formulation is particularly useful for SSMs under weak supervision, where coarse or sparsely sampled data must be lifted to high-resolution reconstructions \cite{sun2024recent}. 
However, most implicit approaches focus primarily on reconstruction and rely on simple or fixed latent priors, without explicitly imposing a structured probabilistic model over the latent space. 
As a result, while they achieve high-fidelity reconstruction, they often lack compact, probabilistically grounded, and interpretable statistical embedding amenable to generative population-level analysis \cite{chen2019learning}.

This work addresses these challenges by augmenting a continuous implicit shape representation with a structured probabilistic latent prior.  
Shapes are represented as neural occupancy fields, allowing evaluation at arbitrary resolution and eliminating the need for explicit correspondences. 
We adopt an autodecoder formulation to directly optimize per-instance latent codes under sparse supervision, decoupling representation learning from density modeling.
To structure the latent space, an autoregressive normalizing flow \cite{huang2018neural} is introduced, replacing the weak isotropic prior with an expressive, tractable density model. 
Additionally, Automatic Relevance Determination (ARD) prior is incorporated to promote compactness through dimension-wise variance regularization \cite{saha2025ard}, encouraging adaptive relevance weighting across latent dimensions. 
Together, these components yield a latent representation that is probabilistically grounded, more compact, and more interpretable, while maintaining high reconstruction quality. 
The resulting framework enables high-resolution reconstruction from sparse and anisotropic inputs without compromising geometric fidelity. It supports downstream analysis via structured latent features and provides a generative prior capable of sampling anatomically plausible shapes. 
By unifying implicit representation learning, probabilistic density estimation, and adaptive latent regularization, the proposed framework establishes a sparse-supervised generative model of anatomical variability.
%
%Statistical analysis of the learned latent space further demonstrates improved representation quality and robustness under weak supervision. 
Thus, the main contributions of this can be outlines as follows:

\begin{itemize}
    \item We propose \name, a sparse-supervised generative shape modelling framework combining neural implicit autodecoders with an autoregressive normalizing flow prior.
    %We propose \name, a sparse-supervised generative shape modeling framework that integrates neural implicit autodecoder-based representations with an autoregressive normalizing flow prior. %This combination yields a structured and expressive latent density capable of modeling population-level anatomical variability. 

    \item We use an ARD-based latent weighting in the flow prior to promote compact representations while preserving generative capacity and reconstruction fidelity. %This enables anatomically plausible sampling and structured analysis of dominant modes of variation.
    
    \item We present comprehensive quantitative evaluation under weak supervision, including uncertainty analysis and high-resolution reconstruction from sparse, anisotropic inputs, showing improved robustness and representation quality.% on public femur and lumbar datasets.

\end{itemize}

%% file: Methods.tex
\section{Methodology}

\begin{figure}[!bht]
    \vspace{-1em}
    \centering
    \includegraphics[trim={0.5cm 1.0cm 10.9cm 6.3cm}, clip=true,width=0.72\linewidth]{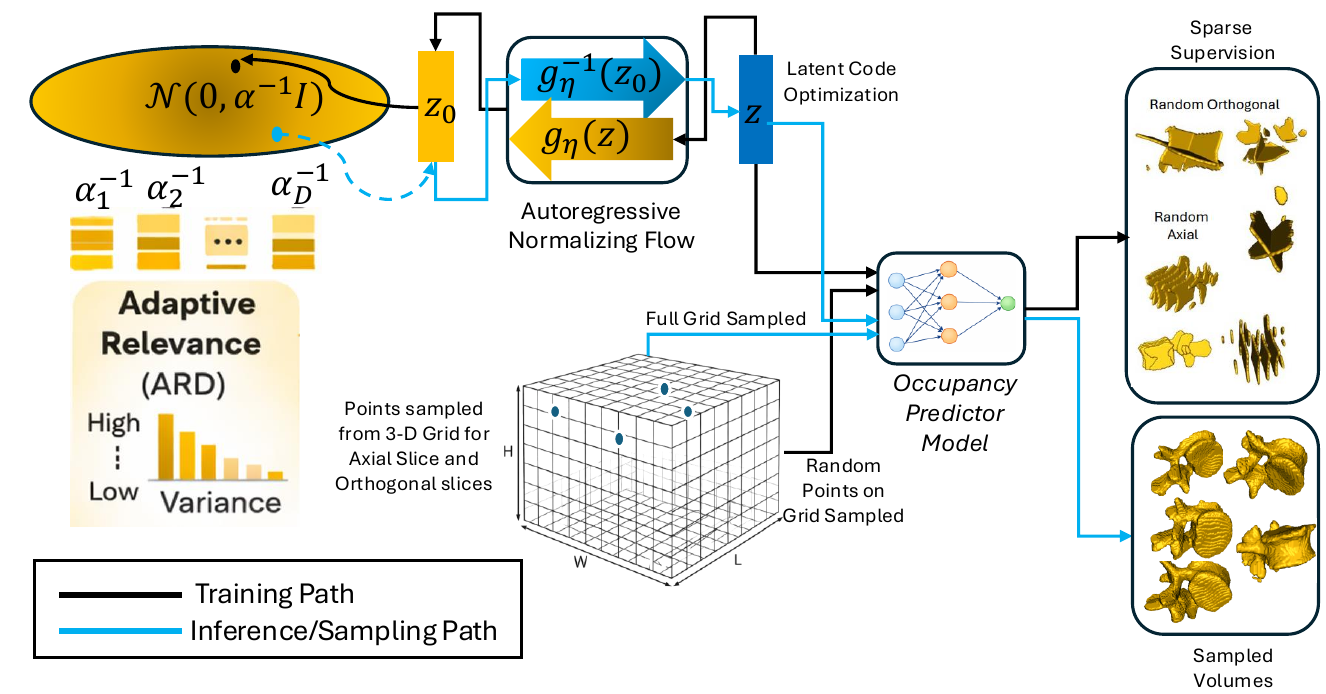}
    \vspace{-1em}
    \caption{\textbf{\name Architecture}. For only generative shape modelling $\alpha^{-1}=1$, making the distribution a standard normal. For compact shape latent $\alpha^{-1}$ values are learned during training with ARD regularization loss.}
    \label{fig:1}
    \vspace{-2em}
\end{figure}

\name (Figure \ref{fig:1}) learns a compact, generative model of anatomical shape variability under sparse supervision. Built on neural implicit autodecoders \cite{park2019deepsdf,amiranashvili2024learning}, it introduces an autoregressive normalizing flow prior for expressive, tractable density modeling and ARD to promote compact, interpretable latent structure. The implicit formulation enables resolution-agnostic reconstruction from sparse inputs for training and inference, while the unified objective jointly optimizes reconstruction, likelihood, and latent compactness.

\vspace{0.05in}
\noindent\textbf{Background: Implicit Shape Representation.} 
Implicit representations model shapes as continuous scalar fields \cite{chen2019learning}, typically parameterized by a multilayer perceptron (MLP) that predicts a continuous occupancy probability field \cite{mescheder2019occupancy,amiranashvili2024learning} over a fixed grid.
Unlike template-based or voxel-based approaches, these models support arbitrary spatial resolutions evaluations \cite{oktay2017anatomically}. This continuous formulation is particularly advantageous in medical imaging, where segmentation masks are often sparse, anisotropic, and acquired at heterogeneous resolutions. 

Each training shape is associated with a latent code $\mathbf{z} \in \mathbb{R}^D$, and a shared decoder network $f_\theta: \mathbb{R}^D \times \mathbb{R}^3 \rightarrow [0,1]$ predicts the occupancy probability at spatial coordinate $\mathbf{x} \in \mathbb{R}^3$: 
$f_\theta(\mathbf{z}, \mathbf{x}) = \hat{y} \in [0,1].$. Given sparse segmentation masks, we sample spatial coordinates and their corresponding occupancy labels to supervise the network. The reconstruction objective is formulated as:
\begin{equation}
\mathcal{L}_{\text{rec}} =
\sum_{(\mathbf{x}, y) \in \mathcal{D}}
\mathrm{Dice}(f_\theta(\mathbf{z}, \mathbf{x}), y),
\end{equation}
where $\mathcal{D}$ denotes the set of sampled spatial points and their corresponding binary occupancy labels.
%This formulation treats shape reconstruction as probabilistic occupancy prediction, enabling stable supervision even under sparse and anisotropic sampling. The latent codes $\mathbf{z}$ are optimized jointly with the decoder parameters $\theta$, yielding a low-dimensional embedding that captures the population-level shape variability under sparse supervision without requiring explicit point correspondences or volumetric grids.
This formulation treats reconstruction as probabilistic occupancy prediction, enabling stable learning from sparse, anisotropic data. The latent codes $\mathbf{z}$ are jointly optimized with the decoder parameters $\theta$, yielding a low-dimensional embedding of population-level shape variability without explicit correspondences or volumetric grids.

% =====================================================
\vspace{0.05in}
\noindent\textbf{Probabilistic Latent Prior via Normalizing Flows.} 

To enable structured probabilistic modeling and principled sampling, we impose an autoregressive normalizing flow prior over the latent space. Unlike the standard isotropic Gaussian, it captures complex dependencies and models population-level anatomical variability through explicit likelihood-based density estimation.

Let $\mathbf{z}$ denote the learned latent code. Rather than directly assuming a simple prior over $\mathbf{z}$, we introduce an invertible mapping $\mathbf{z_0} = g_\eta(\mathbf{z})$, where $g_\eta$ is parameterized as a masked autoregressive rational quadratic spline transform \cite{durkan2019neural}. This transformation maps complex latent distributions to a simple base distribution while preserving exact likelihood evaluation. Unlike coupling-based flows with global mixing, the autoregressive formulation preserves a structured dependency across dimensions without introducing learned linear rotations, maintaining interpretability of the latent axes and enabling dimension-wise relevance analysis.
The base distribution in sampling space is defined as a diagonal Gaussian:
$p(\mathbf{z_0}) = \mathcal{N}(\mathbf{0}, \mathrm{diag}(\boldsymbol{\sigma}^2))$.
Using the change-of-variable formula, the induced log-density over $\mathbf{z}$ is:
% \begin{equation}
% p_\eta(\mathbf{z}) = p(\mathbf{z_0}) 
% \left| \det \frac{\partial g_\eta(\mathbf{z})}{\partial \mathbf{z}} \right|.
% \end{equation}
% Taking logarithms,
\vspace{-0.25em}
\begin{equation}
\log p_\eta(\mathbf{z}) = 
\log p(\mathbf{z_0}) + 
\log \left| \det \frac{\partial g_\eta(\mathbf{z})}{\partial \mathbf{z}} \right|.
\end{equation}
\vspace{-0.5em}

The autoregressive spline transform \cite{durkan2019neural} enables exact likelihood and sampling with greater flexibility than affine flows, supporting joint optimization of reconstruction and density without variational approximations or posterior collapse.

% =====================================================
\vspace{0.05in}
\noindent\textbf{Sparsity-Induced Latent Compactness via ARD.} 
To promote a compact and interpretable latent space, we introduce Automatic Relevance Determination (ARD) by placing a sparsity-inducing prior on the precision parameters. Under sparse supervision, reconstruction signals alone may not sufficiently regularize latent structure, leading to over-parameterized or diffuse embeddings. ARD addresses this by adaptively regulating the contribution of each latent dimension according to its relevance to the underlying anatomical variability.

Let the diagonal covariance of the base distribution be parameterized by $\sigma_i^2 = \alpha_i^{-1}$,
% \begin{equation}
% \sigma_i^2 = \frac{1}{\alpha_i},
% \end{equation}
where $\alpha_i$ denotes the precision of the $i$-th latent dimension.
We impose a Gamma prior on each precision parameter $\alpha_i \sim \mathrm{Gamma}(a,b)$,
% \begin{equation}
% \alpha_i \sim \mathrm{Gamma}(a,b),
% \end{equation}
with shape $a$ and rate $b$.
The negative log-likelihood (ignoring constants) of the Gamma density is
\begin{equation}
-\log p(\alpha_i)
\propto
-(a-1)\log \alpha_i + b \alpha_i.
\end{equation}
Summing across latent dimensions yields the below ARD regularizer that integrates seamlessly with the flow-based prior by regularizing the base distribution parameters in a unified probabilistic framework.

\vspace{-0.25em}
\begin{equation}
\mathcal{L}_{\text{ARD}} =
\sum_{i=1}^{D}
\left[
-(a-1)\log \alpha_i + b \alpha_i
\right].
\end{equation}

\vspace{-0.25em}

Here $a > 1$ and $b < 1$. This hierarchical prior encourages latent dimensions with limited explanatory power to increase precision (collapse variance), effectively suppressing irrelevant variability while retaining dominant modes of anatomical variation. By parameterizing $\alpha_i = \exp(-2 \log \sigma_i)$, gradients propagate smoothly through the variance parameters. Importantly, this adaptive relevance mechanism allows the model to allocate latent capacity dynamically rather than relying on a manually fixed effective dimensionality, yielding a more compact and generatively coherent representation.

% =====================================================
\vspace{0.05in}
\noindent\textbf{Training Objective.}
The final objective combines reconstruction, flow likelihood, and ARD regularization:
\begin{equation}
\mathcal{L}
=
\mathcal{L}_{\text{rec}}
+
\lambda_{\text{flow}}
\left(
- \log p_\eta(\mathbf{z})
\right)
+
\lambda_{\text{ARD}}
\mathcal{L}_{\text{ARD}},
\end{equation}
where
%\begin{itemize}
%\item 
$\mathcal{L}_{\text{rec}}$ is the reconstruction loss,
%\item 
$- \log p_\eta(\mathbf{z})$ encourages probabilistic consistency of the latent representation under the learned flow-based prior,
%\item 
$\mathcal{L}_{\text{ARD}}$ promotes sparsity and compactness,
%\item 
$\lambda_{\text{flow}}$ and $\lambda_{\text{ARD}}$ balance the contributions.
%\end{itemize}
This objective enforces reconstruction fidelity, probabilistic structure, and latent sparsity, resulting in a sparse-supervised generative model of anatomical variability. The model jointly learns a continuous implicit surface representation, an expressive autoregressive prior, and a structured, compact population embedding within a unified likelihood-based framework.

%% file: Results.tex
\section{Results and Discussion}

\textbf{Implementation and Dataset Details.} We use two public datasets: (a) VERSE, containing 230 training and 57 testing lumbar vertebrae (healthy and pathological), and (b) OAI-ZIB, with 254 training and 100 testing distal femur shapes (healthy and osteoarthritic). For both datasets, 20\% of the training data is used for validation. All models are trained on NVIDIA H200 GPUs with a batch size of 6. %To simulate sparse segmentation scenarios, the models are trained under two different settings: (a) axial sparse sampling, where the number of annotations is reduced by a factor of 8 in axial plane for thin and thick volumes, and (b) random orthogonal sparse sampling, where only three random orthogonal slices—one each from the axial, sagittal, and coronal planes—are selected to learn the shape representation.
To simulate sparse segmentation, models are trained under two settings: (a) 8× reduced axial annotations for thin and thick volumes, and (b) random orthogonal sampling using one axial, sagittal, and coronal slice to learn the shape representation. The evaluation strategy follows \cite{amiranashvili2024learning}, i.e optimizing latent vectors using sparse annotations. We report quantitative metrics and qualitative analysis for both the reconstruction accuracy and generative fidelity of \name.

%\textbf{Evaluation Setup.} Under sparse settings (axial or random orthogonal), the occupancy model is frozen and the latent code ($Z$) is optimized using sparse annotations. The optimized $Z$ is then evaluated on a dense 3D grid to reconstruct the surface; for orthogonal sparsity, the largest slice per plane is used at inference. We report average surface distance, Hausdorff distance (HD, HD95), and Dice scores to assess reconstruction. For ARD, we report the number of latent dimensions explaining 95\%, 97\%, and 99\% of variance across latent dimensions.

%For evaluation under sparse segmentation settings (either parallel or orthogonal sparsity), the occupancy model is first frozen, and an optimized latent representation ($Z$) of the shape is computed using the aforementioned loss function. Once the optimized $Z$ is obtained, it is combined with a dense 3D grid and fed into the occupancy model to generate the corresponding surface representation. For orthogonal sparsity at inference, the biggest slices in each plane are used. We report average surface distance, hausdorff distance, hausdorff distance 95 and dice scores of test shapes, as quantitative metrics for establishing shape reconstruction performance. For ARD, we report number of dimensions which explain 95\%, 97\% and 99\% of the variablity in the shape latent space for different sparsity and latent dimensions.   

\textbf{Quantitative Results.} Quantitative results for the VERSE and Femur datasets are presented in Tables \ref{tab:Quantitative:VERSE} and \ref{tab:Quantitative:Femur}, respectively. The findings indicate that incorporating generative capabilities through Flow modeling and an ARD prior does not lead to a significant reduction in shape reconstruction performance across different latent dimensionalities and sparse input settings, including axial thin slices, axial thick slices, and orthogonal slices. In some cases, particularly with a lower latent dimensionality (e.g., 16 dimensions), we observe improved performance for both the VERSE and Femur datasets. For higher latent dimensionalities, performance metrics remain within the standard variation of the metric distribution. The highest accuracy is achieved by models trained on thin slices compared to orthogonal or thick slices, as expected due to the higher resolution of the thin-slice training data.

\begin{table*}[!htb]
\centering
\small
\scalebox{0.79}{
\begin{tabular}{c|c|cccc|cccccc}
\toprule
& & \multicolumn{4}{c|}{\name w/o flow} 
& \multicolumn{6}{c}{\name} \\
& Dim & ASD $\downarrow$ & HSD95 $\downarrow$ & HSD $\downarrow$ & DSC $\uparrow$
& ASD $\downarrow$ & HSD95 $\downarrow$ & HSD $\downarrow$ & DSC $\uparrow$
& Var & R$^2$-Corr \\ 
\hline
\midrule
\multirow{4}{*}{\begin{sideways}\textbf{{Thin}}\end{sideways}} 

&16  & $0.86_{\pm0.25}$ & $3.41_{\pm1.34}$ & $9.86_{\pm3.05}$ & $0.87_{\pm0.03}$
& $0.84_{\pm0.25}$ & $3.41_{\pm1.62}$ & $9.96_{\pm4.00}$ & $0.87_{\pm0.03}$
& 11-12-13 & 0.76-0.87 \\

&32  & $0.69_{\pm0.18}$ & $2.75_{\pm1.01}$ & $8.62_{\pm2.76}$ & $0.89_{\pm0.02}$
& $0.72_{\pm0.25}$ & $3.04_{\pm1.51}$ & $9.16_{\pm3.19}$ & $0.89_{\pm0.02}$
& 20-21-23 & 0.72-0.85 \\

&64  
& $0.57_{\pm0.16}$ & $2.38_{\pm1.07}$ & $8.05_{\pm3.03}$ & $0.91_{\pm0.02}$
& $0.60_{\pm0.23}$ & $2.64_{\pm1.28}$ & $8.53_{\pm3.5}$ & $0.90_{\pm0.03}$
& 39-44-46 & 0.79-0.89 \\

&128 
& $0.49_{\pm0.13}$ & $2.15_{\pm0.69}$ & $7.78_{\pm2.47}$ & $0.92_{\pm0.02}$
& $0.55_{\pm0.23}$ & $2.57_{\pm1.28}$ & $8.49_{\pm3.98}$ & $0.91_{\pm0.02}$
& 69-79-85 & 0.33-0.58 \\

&256 
& $0.73_{\pm0.11}$ & $1.91_{\pm0.55}$ & $7.12_{\pm2.22}$ & $0.93_{\pm0.01}$
& $0.59_{\pm0.27}$ & $2.74_{\pm1.52}$ & $9.17_{\pm3.78}$ & $0.91_{\pm0.03}$
& 164-174-183 & 0.45-0.67 \\
\hline
\midrule
\multirow{4}{*}{\begin{sideways}\textbf{{Thick}}\end{sideways}} 
&16  
& $1.07_{\pm0.24}$   & $4.20_{\pm1.51}$ & $12.10_{\pm3.37}$ & $0.85_{\pm0.03}$
& $0.88_{\pm0.26}$ & $3.41_{\pm1.58}$ & $9.53_{\pm3.70}$ & $0.87_{\pm0.03}$ & 11-12-13 & 0.71-0.84 \\

&32  
& $0.73_{\pm0.15}$ & $2.68_{\pm0.78}$ & $8.30_{\pm2.41}$ & $0.89_{\pm0.02}$
&   $0.74_{\pm0.19}$	 & $2.86_{\pm1.00}$ &	$8.39_{\pm2.98}$ &	$0.89_{\pm0.02}$ &	20-22-24 &	0.73-0.85 \\

& 64& $0.66_{\pm0.15}$ & $2.79_{\pm1.09}$& $8.65_{\pm3.07}$ & $0.90_{\pm0.02}$ & $0.65_{\pm0.25}$ & $2.56_{\pm1.12}$ & $7.85_{\pm2.85}$ & $0.90_{\pm0.03}$ & 38-43-46 & 0.68-0.82\\
&128 
& $0.51_{\pm0.10}$ & $2.04_{\pm0.59}$ & $6.84_{\pm2.24}$ & $0.91_{\pm0.01}$
& $0.60_{\pm0.19}$ & $2.48_{\pm1.11}$ & $8.68_{\pm4.09}$ & $0.91_{\pm0.02}$ & 73-82-89 & 0.36-0.65 \\

&256 
& $0.47_{\pm0.09}$ & $1.84_{\pm0.41}$ & $6.77_{\pm1.98}$ & $0.92_{\pm0.01}$
& $0.62_{\pm0.24}$ & $2.58_{\pm1.28}$ & $8.96_{\pm3.50}$ & $0.90_{\pm0.03}$ & 164-176-185 & 0.52-0.72 \\
\bottomrule
\end{tabular}}
\caption{\textbf{Quantitative Results on VERSE.} Average Surface Distance (\textbf{ASD}), Hausdorff Distance (\textbf{HSD}, \textbf{HSD95}), and Dice Score (\textbf{DSC}) are reported for all models. \textbf{Var} denotes the number of latent dimensions explaining 95–97–99\% variance after ARD regularization. \textbf{R$^2$} and \textbf{Corr} (Pearson r) evaluate predictive performance.}
\label{tab:Quantitative:VERSE}
\vspace{-2em}
\end{table*}

\begin{table*}[!htb]
\centering
\small
\scalebox{0.78}{
\begin{tabular}{c|c|cccc|cccccc}
\toprule
& & \multicolumn{4}{c|}{\name w/o flow} 
& \multicolumn{6}{c}{\name} \\
& Dim & ASD $\downarrow$ & HSD95 $\downarrow$ & HSD $\downarrow$ & DSC $\uparrow$
& ASD $\downarrow$ & HSD95 $\downarrow$ & HSD $\downarrow$ & DSC $\uparrow$
& Var & R$^2$ \\ 
\hline
\midrule
\multirow{4}{*}{\begin{sideways}\textbf{{Thin}}\end{sideways}} 
&16  & $1.82_{\pm0.47}$ & $5.42_{\pm1.40}$ & $13.80_{\pm3.83}$ & $0.94_{\pm0.01}$
& $1.14_{\pm0.25}$ & $3.55_{\pm0.82}$ & $10.65_{\pm3.64}$ & $0.96_{\pm0.01}$ & 11-12-13 & 0.69-0.83 \\

&32  & $1.21_{\pm0.27}$ & $3.79_{\pm0.89}$ & $10.85_{\pm3.02}$ & $0.96_{\pm0.01}$ 
& $0.87_{\pm0.19}$ & $2.82_{\pm0.65}$ & $9.30_{\pm3.08}$ & $0.97_{\pm0.00}$ & 20-22-24 & 0.55-0.74 \\

&64  
& $0.84_{\pm0.17}$ & $2.75_{\pm0.60}$ & $9.13_{\pm3.26}$ & $0.97_{\pm0.00}$
& $0.70_{\pm0.15}$ & $2.35_{\pm0.55}$ & $8.64_{\pm2.87}$ & $0.98_{\pm0.00}$ & 39-44-47 & 0.68-0.82 \\
&128 
& $0.64_{\pm0.12}$ & $2.19_{\pm0.46}$ & $8.14_{\pm2.53}$ & $0.98_{\pm0.00}$
& $0.64_{\pm0.26}$ & $2.18_{\pm0.91}$ & $8.14_{\pm2.95}$ & $0.98_{\pm0.01}$ & 76-85-91 & 0.62-0.79 \\

&256 
& $0.55_{\pm0.10}$ & $1.90_{\pm0.42}$ & $7.39_{\pm2.56}$ & $0.98_{\pm0.00}$ 
& $0.64_{\pm0.35}$ & $2.22_{\pm1.09}$ & $8.24_{\pm3.19}$ & $0.98_{\pm0.01}$ & 163-176-186 & 0.55-0.74 \\
\hline
\midrule
\multirow{4}{*}{\begin{sideways}\textbf{{Thick}}\end{sideways}} 
&16  
&$1.14_{\pm0.25}$ & $3.55_{\pm0.82}$ & $10.65_{\pm3.64}$ & $0.96_{\pm0.01}$
& $1.12_{\pm0.23}$ & $3.49_{\pm0.78}$ & $10.24_{\pm3.08}$ & $0.96_{\pm0.01}$ & 11-12-13 & 0.71-0.84 \\
&32  
& $0.87_{\pm0.17}$ & $2.82_{\pm0.64}$ & $9.30_{\pm2.95}$ & $0.97_{\pm0.00}$ 
& $0.90_{\pm0.19}$ & $2.89_{\pm0.64}$ & $9.21_{\pm2.93}$ & $0.97_{\pm0.00}$ & 20-22-24 & 0.53-0.73 \\

&64  
& $0.61_{\pm0.14}$ & $2.12_{\pm0.54}$ & $7.84_{\pm2.51}$ & $0.98_{\pm0.00}$ 
& $0.72_{\pm0.15}$ & $2.41_{\pm0.53}$ & $8.51_{\pm2.59}$ & $0.98_{\pm0.00}$ & 40-44-48 & 0.64-0.80 \\

&128 
& $0.56_{\pm0.13}$ & $1.90_{\pm0.46}$ & $7.53_{\pm2.32}$ & $0.98_{\pm0.00}$ 
& $0.64_{\pm0.26}$ & $2.18_{\pm0.75}$ & $8.05_{\pm2.57}$ & $0.98_{\pm0.01}$ & 75-84-91 & 0.66-0.817 \\

&256 
& $0.37_{\pm0.11}$ & $1.35_{\pm0.43}$ & $6.70_{\pm2.31}$ & $0.99_{\pm0.00}$ 
& $0.80_{\pm0.61}$ & $2.72_{\pm2.09}$ & $9.38_{\pm5.38}$ & $0.97_{\pm0.02}$ & 166-177-186 & 0.56-0.75 \\
%\multirow{4}{*}{\begin{sideways}\textbf{{Ortho}}\end{sideways}} 
\bottomrule
\end{tabular}}

\caption{\textbf{Quantitative Results for Femur.}}\label{tab:Quantitative:Femur}
\vspace{-4.0em}
\end{table*}

\begin{figure}[!htb]
\vspace{-1em}
    \centering
    \includegraphics[trim={0.0cm 0.0cm 0.0cm 0.cm}, clip=true,width=0.9\linewidth]{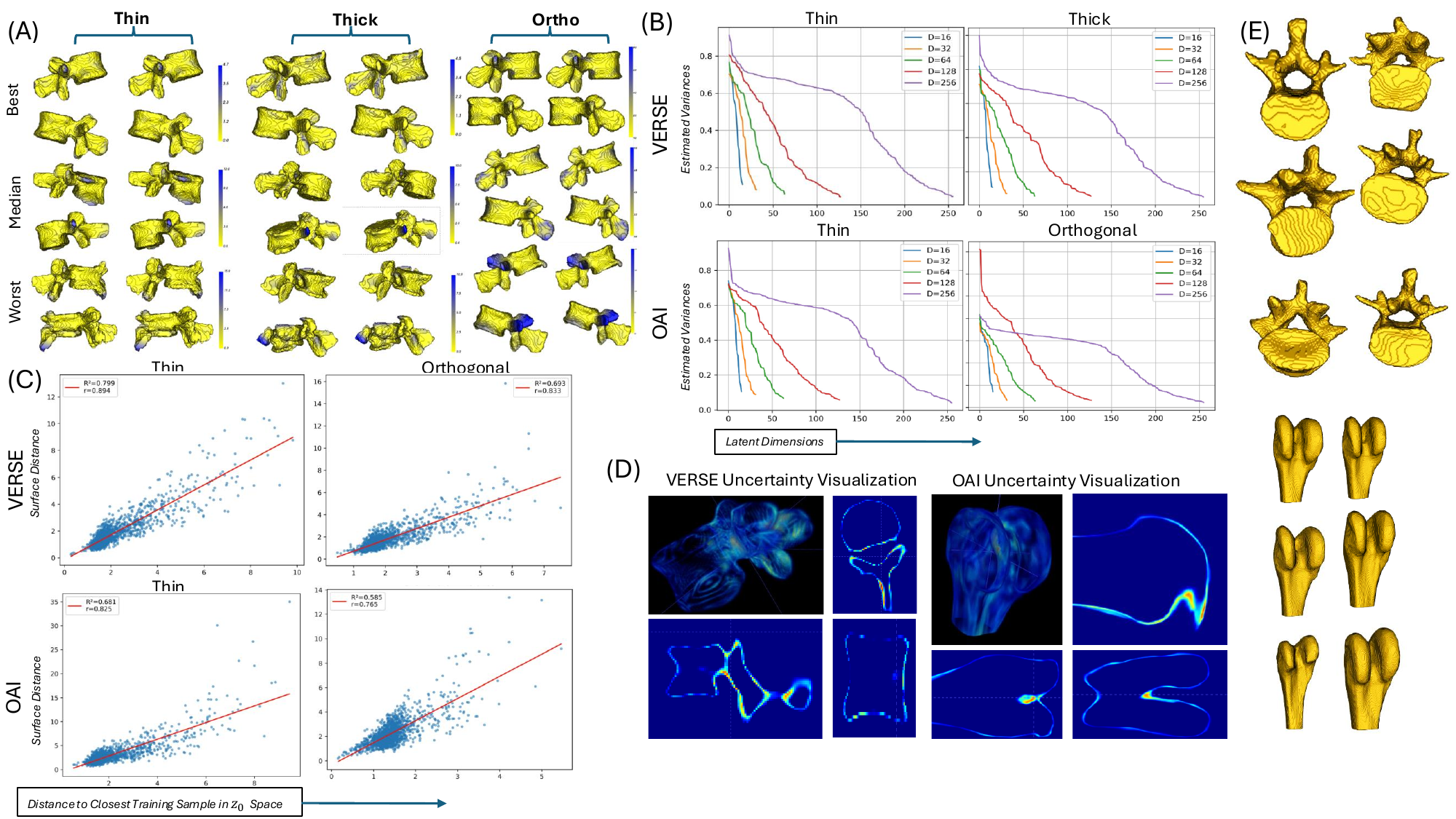}
    \caption{\textbf{Results.} \textbf{(A)} Best, median, and worst surface reconstructions obtained from models trained on thin, thick, and orthogonal slices(latent dimension=64). The right column shows models trained with \name, while the left column shows models trained without Flow and ARD regularization. \textbf{(B)} Estimated $\sigma$ values across different latent dimensionalities when using \name showing the impact of ARD regularization. \textbf{(C)} Correlation plots for 1500 generated samples, showing the relationship between the distance in $z_0$ space to the nearest training sample and the corresponding surface distance to the closest training surface. \textbf{(D)} Uncertainty visualization for one sample from the VERSE and OAI datasets using the best-performing thin-slice model(64 latent dimention), including a 3D rendering of uncertainty and cross-sectional views in the axial, sagittal, and coronal planes.\textbf{(E)} Generated Shapes from VERSE and OAI models trained with latent dimension 64.}
    \label{fig:qualitative_results}
    \vspace{-1.5em}
\end{figure}

\textbf{Qualitative Results.} Qualitative reconstructions for the best, median, and worst cases are shown in Figure~\ref{fig:qualitative_results}A. Models trained with and without Flow-ARD yield comparable reconstruction errors across thin, thick, and orthogonal sparse settings, indicating that the proposed regularization does not degrade accuracy. Orthogonal models exhibit higher errors due to substantially sparser supervision, particularly in geometrically complex regions.

%The results indicate that models trained with and without Flow-ARD exhibit very similar reconstruction errors in comparable regions for thin, think and orthogonal sparse setting. This further demonstrates qualitatively that \name does not result in any reduction in reconstruction accuracy. Higher errors are observed for the orthogonal models, which is expected since they are trained on significantly sparser data, making them more prone to inaccuracies in geometrically complex regions.

\textbf{Learned Sigma Variations.} The learned latent variances are visualized in Figure~\ref{fig:qualitative_results}B by sorting $\sigma$ values in descending order across different latent dimensionalities. The results show that the model concentrates population-level variation within a subset of latent dimensions. ARD regularization consistently promotes compactness, driving approximately 40\% of dimensions toward near-zero variance, thereby yielding a more compact latent representation without degrading reconstruction quality.
%We visualize the variances learned during training by sorting them in descending order for different latent dimensionalities and comparing the $\sigma$ values obtained with ARD priors, as shown in Figure \ref{fig:qualitative_results}B. The results demonstrate that the model distributes the variance of the underlying shape population across a subset of latent dimensions. Notably, ARD regularization consistently promotes sparsity across different latent dimensionalities, driving approximately 40\% of the latent dimensions toward near-zero $\sigma$ values.

\textbf{Evaluating the Generative Capabilities of the Learned Model.}
%To evaluate the generative capabilities of the model and assess whether the generated shapes reflect the underlying training distribution, we sample 1,500 shapes from the learned Gaussian latent distribution. We then compute the correlation between (i) the distance in the $z_0$ latent space to the nearest training sample and (ii) the corresponding surface-to-surface distance between the generated shape and that nearest training shape. A higher correlation indicates that the generated samples more faithfully follow the structure of the training shape population. The resulting correlation plots are shown in Figure \ref{fig:qualitative_results}. We observe a strong correlation for both datasets ($> 0.80$), suggesting that the generated shapes can be reliably regarded as samples from the same underlying distribution as the training data. 
To assess generative fidelity, 1,500 shapes are sampled from the learned latent distribution. For each generated shape, the correlation between (i) its distance in the $z_0$ space to the nearest training sample and (ii) the corresponding surface-to-surface distance is computed.  High correlation ($>0.80$ across datasets; see Figure~\ref{fig:qualitative_results}) indicates that latent distances reflect geometric similarity and that generated samples remain consistent with the training population. This result has important implications for statistical shape modeling approaches that require large cohorts for stable convergence but are often constrained by limited data availability. The proposed framework can generate realistic synthetic shapes that may be leveraged for data augmentation in such data-scarce scenarios. Sample generated shape are shown in Figure \ref{fig:qualitative_results}E. 

\textbf{Uncertainty Quantification.} To improve the interpretability of our results, we additionally perform uncertainty quantification for test shapes by sampling different sparse slice configurations and predicting the corresponding surfaces. Uncertainty is estimated using per pixel entropy from 25 predicted  shapes from different initial conditions of slices and sparsity. The results are presented in Figure \ref{fig:qualitative_results}D. We observe that regions with high curvature exhibit the greatest uncertainty, suggesting that autodecoder-based implicit models are more challenged in geometrically complex areas. This highlights the need for further methodological improvements to reduce uncertainty and enable more reliable surface reconstructions for statistical shape modeling.

